\documentclass[10pt,letterpaper]{article}
\usepackage[utf8]{inputenc}
\usepackage{graphicx}
\usepackage{amsmath,amssymb}
\usepackage{hyperref}
\usepackage{authblk}
\usepackage{booktabs}
\usepackage{multirow}
\usepackage{makecell}

\title{Procedural Memory for Language Agents: \\A Neurosymbolic Perspective on \\Explicit Goal Structure}

\author[1]{Jonathan Leung\thanks{Corresponding Author: jonathan.leung@ntu.edu.sg}}
\affil[1]{Tencent-NTU Joint Research Laboratory (CENTURY), Nanyang Technological University, Singapore}
\author[2]{Zhiqi Shen\thanks{zqshen@ntu.edu.sg}}
\affil[2]{College of Computing and Data Science, Nanyang Technological University, Singapore}
\author[1]{Deheng Ye\thanks{ydyl1991@gmail.com}}

\date{}

\begin{document}

\maketitle

\begin{abstract}
Large language models remain unreliable on long-horizon tasks requiring procedural reasoning. We argue that a key cause is the absence of explicit procedural memory: accessible, modifiable knowledge of how tasks are accomplished, distinct from semantic and episodic memory. Goal-Oriented Graphs offer such a representation, encoding goals and prerequisite dependencies as inspectable symbolic structure. We examine their construction, the challenges they pose, and principles for integrating them with neural reasoning.
\end{abstract}

\noindent\textbf{Keywords:} Procedural memory; Neurosymbolic AI; 
Large language models; Long-horizon planning; Knowledge graphs; Explainable AI; Cognitive architectures

\section*{Introduction}
\label{sec:intro}

Large language models (LLMs) have driven a paradigm shift in artificial intelligence, moving the field from static text processing toward autonomous agentic systems capable of interacting with open-ended, dynamic environments~\cite{wang2024survey,ye2024moreagents,ye2024avalon,ye2024affordable}. By utilizing the LLM as a ``cognitive core,'' these agents interleave reasoning with action by translating high-level linguistic intentions into executable environmental commands through a process known as grounding~\cite{yao2023react}. The result is a new class of systems where language serves as the universal interface between reasoning and control.

However, as agents are deployed in increasingly complex, multi-step settings, long-horizon planning has emerged as a fundamental barrier. By ``long-horizon'', we refer to, in working terms, tasks whose success depends on the correct ordering and satisfaction of prerequisites across a sequence of steps longer than the span over which current LLMs can reliably track dependencies in context. Despite their vast internal knowledge, LLMs frequently fail at plan generation in deterministic environments~\cite{valmeekam2023planbench, valmeekam2023planning}. Crucially, these failures are not primarily knowledge failures. An LLM may know what a diamond pickaxe is, what a deployment pipeline consists of, or what a diagnostic protocol requires, yet still generate invalid action sequences, mistrack prerequisites, or accumulate errors across steps until a plan fails irrecoverably. Evidence of these failure modes has been documented across software engineering~\cite{jimenez2024swe,ye2023rltf}, clinical reasoning~\cite{singhal2023large}, and web navigation~\cite{sodhi2024step}, suggesting this is a general limitation rather than a domain-specific artefact. Position papers have further argued that LLMs cannot plan in isolation and require external structural support~\cite{kambhampati2024position}.

This gap stems partly from a representational deficit: LLMs may possess procedural knowledge implicitly, but they lack an accessible, modifiable, and inspectable procedural representation. Procedural memory encodes knowledge of ``how to do things'': how tasks decompose into subgoals, what prerequisites must be satisfied before each step, and in what order operations must be executed. While LLMs absorb procedural patterns during pretraining~\cite{ruis2025procedural}, this knowledge remains implicit and difficult to inspect, modify, or compose. Classical planning research has long exploited precisely the structure that is missing: hierarchical task networks (HTNs) organize tasks as decompositions with explicit prerequisite structure~\cite{erol1994htn}. Explicit goal structure represents one particularly important symbolic realization of procedural memory and a powerful complement to neural reasoning in long-horizon agentic settings, and its integration constitutes a distinct and under-explored design dimension for language agents. How goal graphs relate to classical task networks such as HTNs is examined in the section on Goal-Oriented Graphs.

We focus on Goal-Oriented Graphs (GoGs)~\cite{LEUNG2026115706} as a symbolic instantiation of procedural memory: directed graphs in which nodes represent goals and edges encode prerequisite dependencies, capturing the hierarchical decomposition of complex tasks in an inspectable and compositionally reusable form. This fits within a neurosymbolic paradigm~\cite{garcez2023neurosymbolic}: the LLM provides subsymbolic capabilities including flexible language understanding and adaptive reasoning, while the GoG instantiates System 2-like scaffolding~\cite{kahneman2011thinking} that constrains the planning space without enforcing rigid execution policies.

The contributions of this perspective are threefold: (1) a conceptual taxonomy of procedural memory representations for language agents, organized by representational locus and trade-off profile; (2) positioning Goal-Oriented Graphs (GoGs) as a principled symbolic instantiation of procedural memory within a neurosymbolic agent architecture; and (3) an argument that explicit goal structure provides enhanced inspectability and supports compositional reuse of task structure, which may facilitate compositional generalization compared to implicit and experience-based alternatives.

Our argument proceeds as follows. In the section on procedural memory in agent architectures, we position procedural memory within the broader agent memory landscape, arguing that it constitutes a distinct and underappreciated component alongside semantic and episodic memory. In the section on Goal-Oriented Graphs, we examine Goal-Oriented Graphs as a concrete symbolic instantiation, grounding the GoG discussion in empirical evidence from Minecraft evaluation~\cite{LEUNG2026115706} and situating it within a broader pattern of procedural reasoning failures documented across software engineering~\cite{jimenez2024swe} and clinical reasoning~\cite{singhal2023large}. We conclude with key challenges, including construction, runtime failure modes, and integration, and discuss design principles for robust goal-directed agency.

\section*{Procedural Memory in Agent Architectures}
\label{sec:memory}

The research landscape for LLM-based agents has coalesced around several complementary memory types, a distinction that mirrors findings from cognitive science. In humans, semantic, episodic, and procedural memory are often treated as distinct memory systems in cognitive science, although consolidation research suggests substantial interaction and gradual transformation between them over time~\cite{tulving1972episodic, mcclelland1995there,frankland2005organization}. This tripartite division is a useful idealization rather than a strict biological taxonomy. Consolidation research suggests that episodic traces can gradually be abstracted into more schema-like procedural knowledge through repeated reactivation~\cite{mcclelland1995there, frankland2005organization}. For LLMs, however, there is no clear analogue of the offline replay and consolidation mechanisms thought to support this transition. As a result, the distinction between what an agent has done (episodes) and what it knows how to do (procedures) remains an operationally useful design distinction, even if the boundary is less sharp in biological systems.
Computational cognitive architectures have long made analogous distinctions, from SOAR's procedural/declarative memory split~\cite{newell1994unified} and ACT-R's production rules~\cite{anderson2013architecture} to Minsky's frame-based representations~\cite{minsky1974framework}. Drawing on cognitive perspectives on agent memory~\cite{sumers2023cognitive}, the literature on LLM agents can be organized around three complementary forms.

\textbf{Semantic memory} (RAG and knowledge graphs~\cite{NEURIPS2020_6b493230, edge2025localglobalgraphrag, pan2024unifying,ye2021relationtransformer}) stores factual knowledge: entities, relations, and domain concepts. It addresses ``what is true'' but does not specify the sequence of operations required to complete a task.

\textbf{Episodic memory} (experience replay and self-reflection~\cite{shinn2023reflexion}) stores records of past interactions, enabling agents to learn from experience. It addresses ``what was done'' but does not encode generalizable procedural knowledge about ``what should be done'' in novel situations. An agent that has successfully completed one multi-step task does not automatically know how to complete a structurally similar but novel task unless it can abstract the underlying procedural dependencies. Some experience-based systems go one step further and extract reusable workflow patterns from recorded trajectories~\cite{wang2025agent,ye2026uivoyager}; such abstractions are boundary cases in the taxonomy proposed here. We classify them as experience-based procedural representations rather than episodic traces: they are derived from episodes, but their function is to encode ``how to do,'' not merely ``what happened,'' and their fidelity is limited by the coverage of the trajectories from which they were distilled.

\textbf{Procedural memory} encodes ``how'' to accomplish tasks. In this paper, procedural memory refers to knowledge of how tasks are accomplished, encompassing action sequences, task decomposition, prerequisite relationships, ordering constraints, and termination conditions. Such knowledge may be represented implicitly in model parameters, distilled from experience as templates or scripts~\cite{fang2025memp, ouyang2025reasoningbank, wang2025agent,ye2023fcdt,ye2022pretrainrl}, or encoded explicitly in symbolic structures such as GoGs. Each representation involves different trade-offs: implicit representations are flexible but opaque; experience-based templates are grounded in real interaction but task-specific; symbolic structures are inspectable and composable but require explicit construction.

Procedural memory encompasses a wide range of knowledge, including motor skills, tool-use routines, control policies, production rules, and goal decompositions. Throughout this perspective, however, we focus on procedural knowledge relevant to language agents operating in symbolic task environments: specifically, structured knowledge of task decomposition and prerequisite relationships, rather than low-level motor policies or reactive control strategies.

\begin{figure}[t!]
    \centering
    \includegraphics[width=\textwidth]{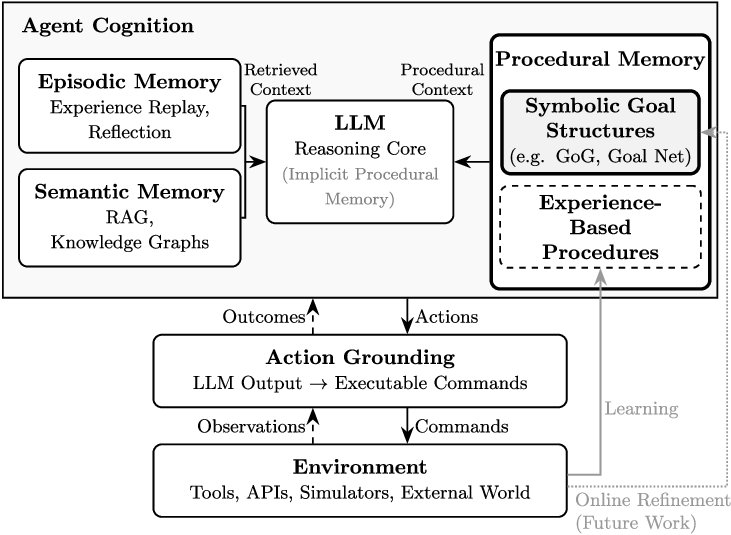}
    \refstepcounter{figure}
    \label{fig:landscape}
\end{figure}

Neither the LLM alone nor any single memory component suffices for reliable long-horizon agency. This limitation has motivated neurosymbolic approaches that combine neural flexibility with explicit symbolic structure~\cite{garcez2023neurosymbolic, manhaeve2018deepproblog, lake2017building}. Some architectures also incorporate world models for simulation-based planning~\cite{ha2018world, hafner2023mastering,ye2024walle,ye2021mapgo,ye2023tfrlsurvey,ye2025crossdomain}, which complement procedural memory by predicting action consequences rather than encoding task structure. This distinction separates reasoning about ``what to do'' from reasoning about ``what may happen given an action sequence.'' Figure~\ref{fig:landscape} illustrates how goal structure relates to these established research directions in agent design.

Importantly, procedural memory need not be exclusively symbolic. Future architectures may combine symbolic goal structures, experience-derived procedural templates, and procedures encoded implicitly within model parameters. The central claim of this perspective is not that all procedural knowledge should be symbolic, but that long-horizon language agents benefit from having at least some procedural structure represented explicitly and available for inspection, retrieval, and modification.

\paragraph{Why existing approaches are insufficient.}
Prompt-based methods such as Chain-of-Thought~\cite{wei2022chain}, Tree of Thoughts~\cite{yao2023tree}, and ReAct~\cite{yao2023react} elicit structured reasoning from the LLM but operate entirely within its latent generative process. They are vulnerable to compounding errors, where a mistake in an early step propagates through subsequent steps with no mechanism for structural correction. Experience-based approaches capture ``what worked before'' rather than ``how to decompose tasks,'' limiting their generalization to novel goal combinations. Fine-tuning embeds procedural knowledge into weights but does not make it inspectable or modular, and fine-tuned knowledge is brittle to distribution shift. Retrieval-augmented methods improve factual accuracy but do not address procedural reasoning. RAG may tell an agent what facts are relevant to a task, while GraphRAG extends this to structured entity relations~\cite{edge2025localglobalgraphrag}, but neither encodes the ordering constraints and prerequisite dependencies required to execute a multi-step plan. Tool-augmented agents partially mitigate execution failures by offloading substeps to external executors such as code interpreters or APIs~\cite{schick2023toolformer, qin2024toolllm,ye2021uismells}, but these approaches improve execution fidelity rather than introducing explicit representations of task decomposition and prerequisite structure.

Given rapid progress in scaling LLMs, a natural question is whether explicit procedural representations are still needed. Scaling context windows increases the volume of information the LLM can attend to but does not impose structural constraints on reasoning. Work on step-by-step verification shows that checking a reasoning trace is a distinct capability from generating it~\cite{lightman2024let}, suggesting that generation alone is insufficient for reliable long-horizon planning. Classical symbolic planners such as STRIPS and PDDL-based systems~\cite{ghallab2004automated, fox2003pddl2} demonstrate that explicit goal representations can support reliable planning when domain models are complete and actions are deterministic, but they require hand-crafted domain specifications that are expensive to construct and unable to handle open-ended natural language. While symbolic representations typically require offline construction, recent work has begun to explore symbolic learning frameworks in which graph-structured agent representations can be updated from experience~\cite{ou2025symbolic}, though GoGs as described here do not yet support online refinement. Scaling and prompting alone have not yet demonstrated a reliable solution to this gap~\cite{valmeekam2023planbench, valmeekam2023planning}, and closing it appears to require a combination of neural flexibility and explicit structure.
\section*{Goal-Oriented Graphs as an Explicit Representation of Procedural Memory}
\label{sec:gog}
Goal-Oriented Graphs externalize procedural structure as an inspectable and modular representation, allowing the agent's reasoning process to draw on explicit task structure rather than relying solely on latent generative planning. It is important to note that GoGs do not exhaust the space of procedural memory. Procedural memory encompasses a broad range of knowledge: motor routines, control policies, tool-use skills, and production rules~\cite{anderson2013architecture, newell1994unified}. GoGs capture one important subset: the decomposition of complex tasks into subgoals with explicit prerequisite and ordering constraints, particularly relevant for language agents operating in domains where tasks can be described in natural language. GoGs are not assumed to be complete or correct symbolic models; instead they are partial, noisy, language-induced structures used as soft constraints on neural planning. Evidence from the Minecraft evaluation~\cite{LEUNG2026115706} suggests that imperfect goal hierarchies reduce task performance, but do not necessarily render planning impossible, indicating a degree of robustness to extraction errors. From a systems perspective, a single GoG simultaneously serves three roles: as a \textit{memory representation} (the stored graph of goals and dependencies), as a \textit{retrieval artifact} (the subgraph extracted for a given task), and as a \textit{planning constraint} (the prerequisite ordering that guides the LLM). These roles are implemented over a single underlying structure that supports storage, lookup, and constraint satisfaction without requiring separate representations. 

Goal-oriented representations are particularly appropriate for language agents because goals are naturally expressed in language and remain stable across variations in execution strategy. A goal such as ``deploy a web service'' or ``diagnose a patient condition'' can be specified, retrieved, and communicated in natural language without requiring a formal domain model. Moreover, multiple action sequences may achieve the same goal depending on available tools, environment state, or agent capability, but the underlying goal decomposition often remains unchanged: deploying a web service requires a passing test suite, a built artifact, and a provisioned environment regardless of which specific tools or scripts the agent uses to accomplish each step. Representing procedural knowledge at the level of goals therefore provides a level of abstraction that is reusable across environments, tools, and execution policies, while remaining interpretable to human users and composable with the LLM's natural language reasoning capabilities. This contrasts with action-level representations such as policies, production rules, or operator schemas, which are typically environment-specific and difficult to inspect or transfer without retraining. This goal-level abstraction is powerful but not universal: it is most effective in domains where procedural knowledge can be reliably extracted and where tasks decompose 
hierarchically, as discussed further in the boundary conditions below.

\begin{figure}[t!]
    \centering
    \includegraphics[width=\textwidth]{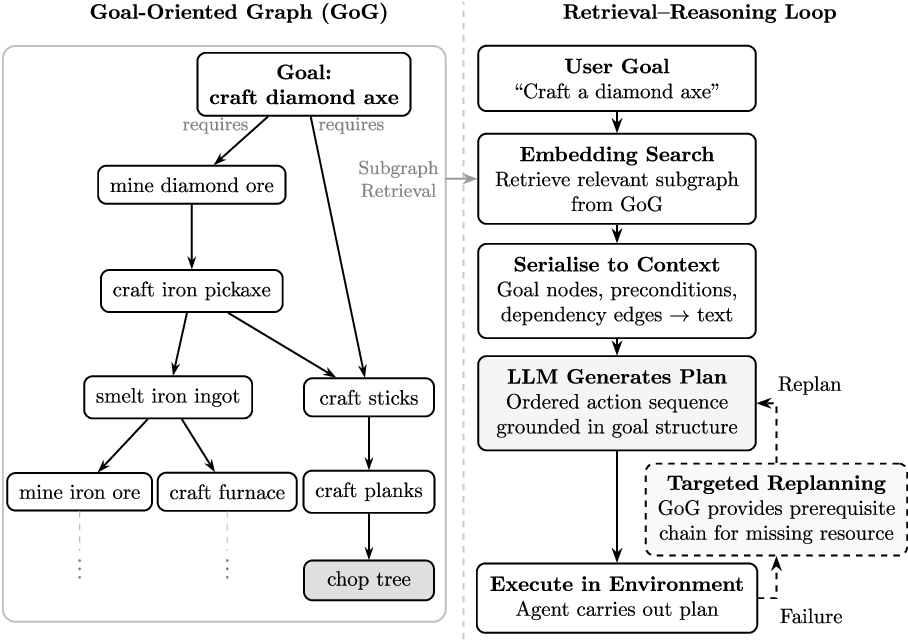}
    \refstepcounter{figure}
    \label{fig:gog}
\end{figure}

\paragraph{The Goal Net formalism and GoG adaptation.}
Goal Net~\cite{shen2006goal, leung2021goal} provides a formalism for representing agent goals as nodes and procedural relationships as directed edges, enabling hierarchical task decomposition. In its original formulation, a learned policy navigates the Goal Net to select which goal to pursue next. Goal-Oriented Graphs adapt this structure for LLM-driven reasoning: instead of a learned goal-selection policy, the LLM traverses the graph to determine which goals to pursue and in what order. Figure~\ref{fig:gog} illustrates the GoG retrieval-reasoning loop~\cite{LEUNG2026115706}. While Goal Nets are typically specified by domain experts, GoGs admit construction from natural-language documentation, making them deployable in domains without formal domain models. In practice, given a user goal, the relevant subgraph is retrieved using embedding-based similarity search. The retrieved subgraph, including goal nodes, preconditions, and dependency edges, is serialised into natural language and injected into the LLM's context. The LLM then reasons over this structured context to produce an ordered plan, grounding abstract goal names into executable actions. If execution fails, the GoG provides the prerequisite chain for the missing resource, enabling targeted replanning without regenerating the entire plan from scratch.

Two aspects of this integration deserve clarification. First, the strength of the constraint is a design choice: in its strongest form, the serialized subgraph is prepended to the LLM context as a hard-constraint block, and each proposed action is checked against the prerequisite ordering of the retrieved subgraph before execution; actions that violate the ordering are rejected or trigger replanning. In practice, integration ranges from a soft prompt to a hard gate of this kind. Second, the graph encodes a partial order: prerequisites that do not depend on one another may be executed in any order, and the LLM may resolve such freedom using task context. Where urgency matters, this ordering should be made explicit in the graph rather than left implicit: in a clinical protocol in which both ``stabilize the patient'' and ``review insurance coverage'' are prerequisites of a surgical goal, the graph either encodes the dependency directly or annotates ``stabilize the patient'' with a priority, so that the safety-critical branch is executed first regardless of retrieval order.

\paragraph{From natural language to GoGs.}
How is a GoG actually constructed? In the Minecraft evaluation, the graph was induced entirely by an LLM: the source corpus (514 Minecraft Wiki pages and 859 in-game recipe files) is split into chunks, an LLM extracts goal nodes with their preconditions and postconditions from each chunk, and the nodes are then merged and linked into a single graph using embedding-similarity checks that decide whether two extracted goals are equivalent, aliases of one another, or related as subgoals~\cite{LEUNG2026115706}. This is an instance of LLM-guided extraction, which we now discuss in general terms. Three construction regimes can be distinguished. (i)~Rule-based extraction from structured documentation (recipe files, manufacturing work instructions) is precise but limited to domains whose procedures are already formally documented. (ii)~LLM-guided extraction, in which an LLM parses natural-language documentation into goal nodes and prerequisite edges, extends coverage to any domain in which procedures are described in prose, but it introduces a bootstrapping problem: an LLM that is prone to ordering errors is used to produce the very structure meant to correct those errors, so an extracted graph must be treated as a hypothesis to be verified, not as ground truth. (iii)~Human-in-the-loop validation, in which domain experts review and correct the extracted graph, remains the strongest available guarantee for high-stakes domains. Practical quality controls combine consistency checks (acyclicity, reachability of atomic goals, cross-checking against source documents), sampled expert review, and updating the graph when execution feedback contradicts its structure, so that runtime failures become maintenance signals. Program induction offers a more ambitious route: systems such as DreamCoder learn libraries of reusable programs directly from data~\cite{ellis2023dreamcoder}, providing an empirically evaluated precedent for the kind of learned procedural library that GoGs currently lack.

\paragraph{Illustrative example.}
Consider the task ``craft a diamond axe'' in Minecraft. A GoG decomposes this goal recursively: a diamond axe requires diamonds and sticks; diamonds require an iron pickaxe in order to be mined; the iron pickaxe requires iron ingots and sticks; iron ingots require iron ore, a furnace, and fuel; and so on until reaching atomic operations. Each goal node encodes preconditions (required tools and materials) and postconditions (produced items), forming a structured chain the LLM can follow rather than generate from scratch. Without this structure, LLMs frequently produce invalid actions (e.g., ``smelt diamond'' instead of ``mine diamond ore'') or miscompute material quantities, causing cascading failures in this setting~\cite{LEUNG2026115706}. Analogous limitations have been documented in software engineering agents~\cite{jimenez2024swe} and clinical reasoning~\cite{singhal2023large}, confirming that the problem is not domain-specific.

\paragraph{The neurosymbolic division of labor.}
GoGs instantiate a specific neurosymbolic integration pattern that can be interpreted through the lens of dual-process theory~\cite{kahneman2011thinking}. For base LLMs without extended reasoning chains, the System~1 characterization is appropriate; extended-reasoning models partially close this gap but do not eliminate prerequisite-tracking failures over long horizons~\cite{valmeekam2023planbench}. The LLM operates as a System 1 reasoner: fast, associative, and pattern-driven, excelling at flexible language understanding but prone to errors in tasks requiring deliberate, step-by-step logical reasoning. The GoG is analogous to System 2 scaffolding: slow, structured, and explicit reasoning about task decomposition and prerequisite ordering. Computationally, this corresponds to explicit constraint satisfaction over a dependency graph, which reduces the effective search space the LLM must explore. The symbolic component provides structural scaffolding that the neural component alone cannot reliably maintain over long horizons; the subsymbolic component provides flexible interpretation of natural language goals and adaptive handling of ambiguity that the symbolic structure cannot supply. Their integration yields a system that is both structurally sound and semantically flexible~\cite{garcez2023neurosymbolic, lake2017building,ye2026wallens}.

\paragraph{Explainability through symbolic structure.}
A consequential property of GoGs is that they make agent reasoning explicitly interpretable. When an LLM agent operates within a GoG, the reasoning trace is not merely a post-hoc rationalization~\cite{turpin2023language} but a traversal of an explicit, auditable goal structure. A human supervisor can inspect the retrieved subgoal chain, verify that prerequisites are correctly identified, and diagnose failures by identifying which dependency was unsatisfied. This is a qualitatively different form of explainability from attention visualization or saliency maps: 
it provides a structural account of why the agent took each action, grounded in an explicit representation of task logic. This property is particularly valuable in high-stakes domains such as clinical decision support~\cite{singhal2023large} or manufacturing, where auditability is a practical requirement. This auditability is, however, conditional: it presumes that the LLM's assessment of which goals are satisfied is itself reliable. If the LLM silently marks a prerequisite as satisfied when it is not, or misclassifies a goal, the traversal remains auditable in form but misleading in substance. Silent misclassification of goal satisfaction is therefore the most dangerous runtime failure mode of a GoG system, and we treat it as a challenge in the Key Challenges section.

\begin{table}[!tb]
    \centering
    \caption{Goal-Oriented Graph structure across representative domains. The fundamental structure 
    is consistent: goals decompose into subgoals with explicit prerequisites. The primary variation is in the source of procedural knowledge available for graph construction. The game domain (Minecraft) is the only one that has been empirically evaluated~\cite{LEUNG2026115706}; the other rows are conceptual illustrations of the representation pattern.}
    \label{tab:domain_generalization}
    \begin{tabular}{p{2.0cm} p{3.3cm} p{2.6cm} p{2.6cm}}
        \toprule
        \textbf{Domain} & \textbf{Example Goal} & 
        \textbf{Prerequisite Structure} & 
        \textbf{Knowledge Source} \\
        \midrule
        Game & Craft diamond axe & 
        Tool tiers, material chains & 
        Wiki pages, recipe files \\
        Software & Deploy web service & 
        Dependency ordering, build artifacts & 
        Documentation, runbooks \\
        Medicine & Diagnose condition X & 
        Prior test results, clinical criteria & 
        Clinical guidelines \\
        Manufacturing & Assemble device Y & 
        Component availability, tool requirements & 
        Technical manuals \\
        \bottomrule
    \end{tabular}
\end{table}

\paragraph{Generalization across domains.}
The structure captured by GoGs is not domain-specific. Table~\ref{tab:domain_generalization} summarizes this across four representative domains. In software engineering, deploying a service requires a passing test suite, built artifact, and provisioned environment, all of which are prerequisites that must be satisfied in order. In medicine, clinical guidelines frequently contain hierarchical structure: administering a treatment typically requires a confirmed diagnosis, absence of contraindications, and established dosing. In manufacturing, assembly sequences have strict component and tool ordering enforced by technical manuals. GoGs are best understood as a representation pattern: wherever tasks decompose into ordered subgoals with explicit prerequisites and procedural knowledge is available in documented form, GoGs provide a principled representation.

\paragraph{A localist representation with delegated grounding.}
The choice to represent procedural knowledge as an explicit, localist structure places GoGs on one side of a long-standing debate in the connectionist literature. Localist (symbolic) representations are inspectable and support systematic composition, but are brittle and require explicit construction; distributed representations are robust and data-driven, but opaque~\cite{hinton1986distributed, smolensky1990tensor}. Intermediate proposals such as sparse distributed~\cite{kanerva1988sparse} and holographic reduced representations~\cite{plate1995hrr} suggest that inspectability and distribution need not be mutually exclusive, and may be relevant to the internalization question raised in the Conclusion. However, one consequence of the localist choice is that the grounding of GoG nodes is not direct but delegated to the LLM's distributed internal knowledge. This delegation inherits the symbol grounding problem~\cite{harnad1990symbol}: the graph makes the structure of a procedure inspectable, but whether each node denotes what it should in the current environment depends on the LLM's semantics, which cannot be verified from the graph alone. Inspectability of structure and correctness of grounding are therefore distinct properties, and the former does not entail the latter.

\paragraph{GoG versus related methods.}
GoGs differ from Graph of Thoughts~\cite{besta2024graph}, 
which models the flow of information between cognitive operations rather than dependencies between task objectives. They differ from classical PDDL-based planners~\cite{ghallab2004automated, fox2003pddl2} and Hierarchical Task Networks (HTNs)~\cite{erol1994htn}, which require hand-crafted domain models specifying all tasks, methods, and operators in a formal language, and which assume the model is complete and correct: a plan is valid only relative to the model. Unlike HTNs and Goal Nets, GoGs are not assumed to be complete or correct symbolic models; instead they are partial, noisy, language-induced structures used as soft constraints on neural planning, and have been shown to have some tolerance to imperfections in the extracted goal hierarchy~\cite{LEUNG2026115706}. Unlike action-level representations such as production rules~\cite{newell1994unified, anderson2013architecture} or PDDL operators, which encode procedural knowledge as condition-action pairs tied to specific execution contexts, GoGs represent knowledge at the level of goals and their dependencies, enabling reuse across different execution strategies without requiring a complete formal domain model.

GoGs also differ from graph-based knowledge retrieval systems. In contrast to entity-centric knowledge graphs used by GraphRAG, which fragment procedural knowledge into low-granularity entity-relation triples, GoGs represent procedural knowledge compactly as goal hierarchies. A GoG constructed from the same source material as GraphRAG contained 703 nodes and 1,653 edges compared to GraphRAG's 12,388 nodes and 18,347 edges (roughly one-seventh and one-tenth of the scale) yet showed improved performance on multi-step planning tasks in a controlled Minecraft evaluation~\cite{LEUNG2026115706}. Across 66 tasks spanning seven difficulty tiers (30 independent runs per task), GoG achieved success rates of 72.3\% and 66.1\% on the two hardest tiers (gold and diamond crafting) with Llama 3.2-Vision (and 72.2\% and 31.7\% with Gemma 3) whereas the best-performing baseline reached at most 7.5\% and 11.5\% on those tiers, respectively; on the diamond tier with Llama 3.2-Vision, every baseline scored 0\%. On a mid-difficulty tier (stone), GoG more than doubled GraphRAG's success rate (80.0\% vs. 37.8\% with Llama 3.2-Vision), and GraphRAG was in some cases no better than no retrieval at all. Other Minecraft agents such as Ghost in the Minecraft~\cite{zhu2023ghostminecraftgenerallycapable,ye2021juewumc,ye2025vistawise,ye2025causalmace,ye2022minerl} and ODYSSEY~\cite{liu2025odyssey} achieve competent long-horizon behavior using text-based knowledge retrieval and skill libraries respectively, but leave procedural planning to the LLM's latent reasoning without explicit goal structure. As a result, the procedural dependencies underlying generated plans are less directly inspectable, and the mechanisms supporting compositional reuse of task structure are less explicit. Optimus-1~\cite{li2024optimus} similarly employs a Hierarchical Directed Knowledge Graph for procedural planning in Minecraft, demonstrating that explicit goal structure improves long-horizon task performance independently of GoG. However, its knowledge graph is constructed solely from recipe files and does not update from interaction experience; multimodal experience is instead stored in a separate experience pool used for in-context learning~\cite{li2024optimus}. This separation supports the future direction discussed in the Key Challenges section: enabling symbolic goal structures to be refined through experience remains an open problem. Recent hybrid systems combine LLMs with hierarchical or constraint-based planners, but typically treat symbolic structure as an execution-time constraint rather than a persistent representational memory system~\cite{kambhampati2024position}.

\paragraph{Boundary conditions.}
Explicit goal structure is most beneficial when three conditions hold: procedural knowledge can be extracted from documentation with reasonable fidelity; tasks decompose hierarchically into goals with identifiable prerequisites; and task success depends on correct ordering and prerequisite satisfaction. GoGs are less suited to fundamentally open-ended or creative tasks where goal hierarchies cannot be specified in advance, tasks involving highly implicit dependencies such as social reasoning, and tasks where goal structure must be discovered through exploration. In domains where tasks are loosely structured, goals are ambiguous, or procedural knowledge is better captured by distributional patterns than explicit hierarchies, purely neural or experience-based approaches may outperform explicit goal structures. Even within suitable domains, GoGs may introduce brittleness when the retrieved decomposition is incomplete or misaligned with the current environment state, leading to structured but incorrect plans. This is distinct from the unstructured failures of purely generative approaches: GoG failures are typically diagnosable through inspection of the 
retrieved subgraph, whereas neural planning failures 
may be difficult to attribute. In the Minecraft study~\cite{LEUNG2026115706}, increasing noise in graph construction reduced task performance but did not eliminate the utility of the retrieved goal structure, suggesting some tolerance to imperfect procedural extraction.
How recovery should be handled when the graph itself is incorrect remains an open question. Potential approaches include execution-driven inconsistency detection, local fallback planning, and offline graph revision; we discuss these challenges further in the Key Challenges section.

\section*{Key Challenges}
\label{sec:challenges}

The neurosymbolic integration of procedural memory with LLM reasoning addresses a genuine gap in agent design, but several challenges remain.

\paragraph{Construction and maintenance.}
GoGs are currently constructed offline from documentation or demonstration data. In domains where procedural knowledge is well-documented and stable, this is manageable. In open-ended or evolving domains, automated extraction must handle ambiguity, conflicting sources, and incomplete documentation. This is not a new problem: the expert systems literature of the 1980s and 1990s identified essentially the same difficulty as the ``knowledge acquisition bottleneck'', which is the cost and fragility of eliciting, formalizing, and maintaining domain knowledge as explicit representations, which was among the principal reasons that paradigm failed to scale~\cite{hayesroth1983expert,cullen1988reassessment}. The lessons of that era inform GoG construction: target domains in which procedures are already documented in writing; combine automated extraction with human validation; and design representations that tolerate incompleteness rather than demanding complete domain models. Trajectory-based construction, which infers goal structures from successful demonstrations, offers a promising alternative for environments such as WebShop~\cite{yao2022webshop} and ALFWorld~\cite{shridhar2021alfworld,ye2021tikick}. It should be noted, however, that this is itself a hard problem: recovering the goal structure that led to successful trajectories is an inverse planning problem, and under partial observability where preconditions are not all directly observable, it inherits the difficulty of planning and inference in partially observable stochastic domains~\cite{kaelbling1998planning}.

\paragraph{Expressiveness and GoG refinement.}
The expressiveness of symbolic procedural representations is often limited to directed acyclic graphs, which cannot represent cyclic dependencies or dynamically evolving goal structures. Whether the acyclicity restriction is fundamental to GoGs or a contingent design choice is an open question: current GoGs are acyclic because recursive prerequisite retrieval is well-defined on DAGs and because cycles complicate the semantics of goal satisfaction. Many real procedural domains, however, contain iterative dependencies that a DAG cannot express, such as debug--fix--retest loops in software engineering, dose titration in clinical protocols, or iterative sprints in project management. Extensions such as AND--OR graphs with shared subgoals, or Petri-net-style constructs in which cycles are first-class objects, are natural directions for increasing the formalism's expressiveness. A related expressiveness gap is the treatment of prerequisites as deterministic. Many procedural dependencies are only probabilistically satisfied, where a test may be required ``if symptoms persist,'' a step may be conditionally skipped. Probabilistic logic programming, such as ProbLog, DeepProbLog~\cite{manhaeve2018deepproblog}, and Markov Logic Networks~\cite{richardson2006markov}, addresses precisely this setting: logical constraints with associated probabilities, and inference that tolerates violations. Incorporating such probabilistic edges into GoGs would provide a principled extension for the stochastic prerequisite-satisfaction problem, rather than the current all-or-nothing treatment. More fundamentally, GoGs are currently constructed offline and do not update from interaction experience. While experience-based procedural representations already learn incrementally from interaction~\cite{wang2025agent,fang2025memp}, and symbolic learning frameworks have demonstrated that agent-level symbolic structures can be iteratively updated from experience~\cite{ou2025symbolic}, GoGs as described here do not yet support online refinement. Enabling GoGs to incorporate interaction experience, discovering more efficient subgoal orderings, pruning unreachable branches, and incorporating newly learned procedures, remains an open direction.

\paragraph{Runtime failure modes.}
The challenges so far concern construction and expressiveness; GoG systems can also fail at runtime, and some of those failures are insidious. Such failures may occur when the GoG itself is wrong, for example due to a missing or inverted dependency, an outdated prerequisite, or an unreachable goal.
While systematic recovery mechanisms have not yet been explored empirically, several plausible strategies exist. First, execution feedback may reveal inconsistencies between the retrieved subgraph and the environment. Second, the agent may temporarily bypass the affected portion of the graph and rely on unconstrained LLM reasoning for that subtask while preserving the remainder of the structure. Third, discrepancies can be logged for offline graph revision.
However, the most dangerous failure is silent goal misclassification: the LLM marks a prerequisite as satisfied when it is not, or fails to recognize that a goal has been achieved. Because the GoG's structure is auditable, an auditor sees a clean, coherent traversal while the underlying assessment is wrong; the trace is a faithful record of what the agent \textit{believed}, not of what was true, so auditability of form does not guarantee auditability of substance. Such failures are likeliest for goals whose satisfaction cannot be verified from context (e.g., ``the environment is safe''), for ambiguously worded goals, and during long subgraph traversals in which earlier assessments are no longer in the model's attention window. Mitigations include requiring environmental evidence (observed postconditions) rather than LLM assertion for high-stakes nodes; confidence thresholds below which the system defers to a human; and probabilistic edges, which make uncertainty explicit in the structure instead of absorbing it into a binary satisfied/unsatisfied label.


\paragraph{Integration and representation tradeoffs.}
Procedural memory must work alongside semantic memory, episodic memory, and world models, not replace them. A concrete starting point for the integration problem is a minimal conflict-resolution policy: the GoG's prerequisite ordering takes precedence over the LLM's own sequencing of actions (a structural hard constraint), while the assessment of whether an individual goal is satisfied is delegated to the LLM, with a confidence threshold below which the system requests environmental evidence or defers to a human. Even so, how the LLM should balance guidance from procedural memory against retrieved factual knowledge, past experiences, and predicted environment dynamics remains an open design challenge~\cite{sumers2023cognitive}. Table~\ref{tab:tradeoffs} summarizes the expected tradeoffs among representation paradigms. The low adaptability of symbolic representations and high adaptability of experience-based ones in this table reflect the stability--plasticity tradeoff in neural learning~\cite{mcclelland1995there, kirkpatrick2017overcoming}: a representation that generalizes by composition must sacrifice the ability to update itself from experience. The complementary-learning-systems view, in which a fast, experience-driven component gradually distills knowledge into a slow, structured store~\cite{mcclelland1995there}, suggests a concrete architectural direction for closing the online-refinement gap: a GoG that is updated by a fast experience-based component rather than learned directly from raw interaction. Symbolic representations achieve inspectability because their semantics are defined explicitly by structure rather than distributed across parameters; compositional generalization is expected to benefit from the same property, just as a novel sentence can be understood by composing known words and grammatical rules~\cite{lake2017building}.

\begin{table}[!tb]
    \centering
    \caption{Expected tradeoffs among procedural memory representations for LLM agents. A ``$\sim$'' indicates partial or moderate support. These ratings are conceptual expectations based on representational properties rather than a quantitative benchmark comparison.}
    \label{tab:tradeoffs}
    \resizebox{\textwidth}{!}{\begin{tabular}{lccc}
        \toprule
        \textbf{Dimension} & \textbf{Implicit} & 
        \textbf{Experience-based} & \textbf{Symbolic} \\
        & \textit{(fine-tuning)} & 
        \textit{(templates)} & \textit{(GoG, PDDL)} \\
        \midrule
        Inspectability & $\times$ & $\sim$ & $\checkmark$ \\
        Compositional generalization & $\sim$ & $\sim$ & 
        $\checkmark$ \\
        Construction cost & High & Medium & Variable \\
        Adaptability & Low & High & Low--Medium \\
        Cross-domain transfer & $\sim$ & $\sim$ & 
        $\sim$ \\
        Error diagnosis & $\times$ & $\sim$ & $\checkmark$ \\
        \bottomrule
    \end{tabular}}
\end{table}

\section*{Conclusions}
\label{sec:conclusion}

LLMs are powerful generative models, but their latent reasoning capabilities are often insufficient for reliable long-horizon agency. The absence of accessible, modifiable, and inspectable procedural memory, a first-class representation of ``how to do things'' distinct from ``what is true'' and ``what has happened'', represents a design gap for which scaling and prompting alone have not yet demonstrated a reliable solution~\cite{valmeekam2023planbench, valmeekam2023planning}.

Before stating design principles, it is worth distinguishing what this paper has demonstrated from what it has argued. The empirical evidence to date covers a single domain: GoG retrieval improves multi-step task performance in Minecraft~\cite{LEUNG2026115706}, and a GoG trace is architecturally distinct from post-hoc rationalization in that it is a traversal of an explicit graph rather than a generated narrative~\cite{turpin2023language}. Cross-domain generalizability, the compositional advantage of goal-level abstraction, and the tradeoffs summarized in Table~\ref{tab:tradeoffs} are arguments from representational properties, not multi-domain demonstrations; as a perspective, this paper should be read accordingly.

Our analysis points to the following design principles for building robust, goal-directed LLM agents:

\begin{enumerate}
    \item \textbf{Procedural knowledge benefits from explicit goal-oriented representation.} The decomposition of goals into subgoals, with explicit prerequisites and ordering constraints, is a distinct form of knowledge that should not be conflated with semantic facts or episodic traces. Representing procedural knowledge explicitly makes it inspectable, modular, and reusable across tasks and environments, provided that the representation is also grounded, since an inspectable but misgrounded graph is no better than a fluent but unfaithful explanation.

    \item \textbf{Because GoGs currently lack online refinement, hybrid architectures are the most tractable near-term direction.} The gap between symbolic structure and experience-based updating identified in the Key Challenges section is precisely the gap that complementary learning systems are designed to exploit~\cite{mcclelland1995there}: a fast, experience-driven component (trajectory-based or template-based) distilling updates into a slow, structured GoG store would combine the inspectability of the symbolic with the adaptability of the experience-based, and is a more promising near-term target than making either component do both jobs.

    \item \textbf{Explicit procedural representations provide a practical path toward auditability, conditional on reliable satisfaction assessment.} Symbolic procedural representations make agent reasoning auditable in form: when a plan fails, the goal structure provides a structural account of which prerequisite was unsatisfied and why. This auditability is real but conditional on the reliability of the LLM's goal-satisfaction assessment discussed in the Key Challenges section; in high-stakes deployments it must be reinforced by environmental verification of postconditions. In the domain where the approach has been demonstrated, GoG traversal provides such a trace; extending the guarantee to new domains requires validating both the structure and the assessments on which it relies.
\end{enumerate}

Procedural memory is not the only missing component in language agent design, but it is among the most consequential and underexplored. The choice of procedural representation, whether symbolic graphs, experience-based templates, or hybrid approaches, will shape the trajectory of language agent reliability in the years ahead. We hope this perspective contributes to making that choice a deliberate one. A remaining question is whether procedural structure should remain an external memory system or eventually be internalized as a learnable, editable representation within foundation models themselves; for the latter, the localist--distributed literature suggests intermediate targets, such as sparse distributed or holographic representations~\cite{kanerva1988sparse, plate1995hrr}, that retain inspectability while residing in parameter space. This choice will shape the long-term trajectory of neurosymbolic agent design.

\section*{Acknowledgements}
This work was supported by Tencent and the Tencent-NTU Joint Research Laboratory (CENTURY), Nanyang Technological University, Singapore.

\section*{Funding}
This work was supported by Tencent and the Tencent-NTU Joint Research Laboratory (CENTURY), Nanyang Technological University, Singapore. The funder supported the research reported above but played no role in study design, analysis, or manuscript preparation.

\section*{Competing Interests}
The authors declare no competing financial or non-financial interests.

\bibliographystyle{naturemag}
\bibliography{ref}

@inproceedings{leung2021goal,
  title={Goal modelling for deep reinforcement learning agents},
  author={Leung, Jonathan and Shen, Zhiqi and Zeng, Zhiwei and Miao, Chunyan},
  booktitle={Joint European Conference on Machine Learning and Knowledge Discovery in Databases},
  pages={271--286},
  year={2021},
  organization={Springer}
}

@misc{edge2025localglobalgraphrag,
    title={From Local to Global: A Graph RAG Approach to Query-Focused Summarization}, 
    author={Darren Edge and Ha Trinh and Newman Cheng and Joshua Bradley and Alex Chao and Apurva Mody and Steven Truitt and Dasha Metropolitansky and Robert Osazuwa Ness and Jonathan Larson},
    year={2024},
    eprint={2404.16130},
    archivePrefix={arXiv},
    primaryClass={cs.CL},
    url={https://arxiv.org/abs/2404.16130}, 
}

@inproceedings{NEURIPS2020_6b493230,
 author = {Lewis, Patrick and Perez, Ethan and Piktus, Aleksandra and Petroni, Fabio and Karpukhin, Vladimir and Goyal, Naman and K\"{u}ttler, Heinrich and Lewis, Mike and Yih, Wen-tau and Rockt\"{a}schel, Tim and Riedel, Sebastian and Kiela, Douwe},
 booktitle = {Advances in Neural Information Processing Systems},
 editor = {H. Larochelle and M. Ranzato and R. Hadsell and M.F. Balcan and H. Lin},
 pages = {9459--9474},
 publisher = {Curran Associates, Inc.},
 title = {Retrieval-Augmented Generation for Knowledge-Intensive NLP Tasks},
 url = {https://proceedings.neurips.cc/paper_files/paper/2020/file/6b493230205f780e1bc26945df7481e5-Paper.pdf},
 volume = {33},
 year = {2020}
}

@inproceedings{wei2022chain,
    author = {Wei, Jason and Wang, Xuezhi and Schuurmans, Dale and Bosma, Maarten and ichter, brian and Xia, Fei and Chi, Ed and Le, Quoc V and Zhou, Denny},
    booktitle = {Advances in Neural Information Processing Systems},
    editor = {S. Koyejo and S. Mohamed and A. Agarwal and D. Belgrave and K. Cho and A. Oh},
    pages = {24824--24837},
    publisher = {Curran Associates, Inc.},
    title = {Chain-of-Thought Prompting Elicits Reasoning in Large Language Models},
    url = {https://proceedings.neurips.cc/paper_files/paper/2022/file/9d5609613524ecf4f15af0f7b31abca4-Paper-Conference.pdf},
    volume = {35},
    year = {2022}
}

@inproceedings{yao2023react,
    title={ReAct: Synergizing Reasoning and Acting in Language Models},
    author={Shunyu Yao and Jeffrey Zhao and Dian Yu and Nan Du and Izhak Shafran and Karthik R Narasimhan and Yuan Cao},
    booktitle={The Eleventh International Conference on Learning Representations },
    year={2023},
    url={https://openreview.net/forum?id=WE_vluYUL-X}
}

@inproceedings{yao2023tree,
 author = {Yao, Shunyu and Yu, Dian and Zhao, Jeffrey and Shafran, Izhak and Griffiths, Tom and Cao, Yuan and Narasimhan, Karthik},
 booktitle = {Advances in Neural Information Processing Systems},
 editor = {A. Oh and T. Naumann and A. Globerson and K. Saenko and M. Hardt and S. Levine},
 pages = {11809--11822},
 publisher = {Curran Associates, Inc.},
 title = {Tree of Thoughts: Deliberate Problem Solving with Large Language Models},
 url = {https://proceedings.neurips.cc/paper_files/paper/2023/file/271db9922b8d1f4dd7aaef84ed5ac703-Paper-Conference.pdf},
 volume = {36},
 year = {2023}
}

@article{valmeekam2023planbench,
  title={Planbench: An extensible benchmark for evaluating large language models on planning and reasoning about change},
  author={Valmeekam, Karthik and Marquez, Matthew and Olmo, Alberto and Sreedharan, Sarath and Kambhampati, Subbarao},
  journal={Advances in Neural Information Processing Systems},
  volume={36},
  pages={38975--38987},
  year={2023}
}

@article{turpin2023language,
  title={Language models don't always say what they think: Unfaithful explanations in chain-of-thought prompting},
  author={Turpin, Miles and Michael, Julian and Perez, Ethan and Bowman, Samuel},
  journal={Advances in Neural Information Processing Systems},
  volume={36},
  pages={74952--74965},
  year={2023}
}

@article{shen2006goal,
  title={Goal-oriented methodology for agent system development},
  author={Shen, Zhiqi and Miao, Chunyan and Gay, Robert and Li, Dongtao},
  journal={IEICE TRANSACTIONS on Information and Systems},
  volume={89},
  number={4},
  pages={1413--1420},
  year={2006},
  publisher={The Institute of Electronics, Information and Communication Engineers}
}

@article{valmeekam2023planning,
  title={On the planning abilities of large language models-a critical investigation},
  author={Valmeekam, Karthik and Marquez, Matthew and Sreedharan, Sarath and Kambhampati, Subbarao},
  journal={Advances in Neural Information Processing Systems},
  volume={36},
  pages={75993--76005},
  year={2023}
}

@inproceedings{shinn2023reflexion,
 author = {Shinn, Noah and Cassano, Federico and Gopinath, Ashwin and Narasimhan, Karthik and Yao, Shunyu},
 booktitle = {Advances in Neural Information Processing Systems},
 editor = {A. Oh and T. Naumann and A. Globerson and K. Saenko and M. Hardt and S. Levine},
 pages = {8634--8652},
 publisher = {Curran Associates, Inc.},
 title = {Reflexion: language agents with verbal reinforcement learning},
 url = {https://proceedings.neurips.cc/paper_files/paper/2023/file/1b44b878bb782e6954cd888628510e90-Paper-Conference.pdf},
 volume = {36},
 year = {2023}
}

@misc{zhu2023ghostminecraftgenerallycapable,
    title={Ghost in the Minecraft: Generally Capable Agents for Open-World Environments via Large Language Models with Text-based Knowledge and Memory}, 
    author={Xizhou Zhu and Yuntao Chen and Hao Tian and Chenxin Tao and Weijie Su and Chenyu Yang and Gao Huang and Bin Li and Lewei Lu and Xiaogang Wang and Yu Qiao and Zhaoxiang Zhang and Jifeng Dai},
    year={2023},
    eprint={2305.17144},
    archivePrefix={arXiv},
    primaryClass={cs.AI},
    url={https://arxiv.org/abs/2305.17144}, 
}

@article{sumers2023cognitive,
  title={Cognitive architectures for language agents},
  author={Sumers, Theodore and Yao, Shunyu and Narasimhan, Karthik R and Griffiths, Thomas L},
  journal={Transactions on Machine Learning Research},
  year={2024}
}

@article{wang2024survey,
  title={A survey on large language model based autonomous agents},
  author={Wang, Lei and Ma, Chen and Feng, Xueyang and Zhang, Zeyu and Yang, Hao and Zhang, Jingsen and Chen, Zhiyuan and Tang, Jiakai and Chen, Xu and Lin, Yankai and others},
  journal={Frontiers of Computer Science},
  volume={18},
  number={6},
  pages={186345},
  year={2024},
  publisher={Springer}
}

@inproceedings{
  kambhampati2024position,
  title={Position: {LLM}s Can{\textquoteright}t Plan, But Can Help Planning in {LLM}-Modulo Frameworks},
  author={Subbarao Kambhampati and Karthik Valmeekam and Lin Guan and Mudit Verma and Kaya Stechly and Siddhant Bhambri and Lucas Paul Saldyt and Anil B Murthy},
  booktitle={Forty-first International Conference on Machine Learning},
  year={2024},
  url={https://openreview.net/forum?id=Th8JPEmH4z}
}

@inproceedings{
wang2025agent,
title={Agent Workflow Memory},
author={Zora Zhiruo Wang and Jiayuan Mao and Daniel Fried and Graham Neubig},
booktitle={Forty-second International Conference on Machine Learning},
year={2025},
url={https://openreview.net/forum?id=NTAhi2JEEE}
}

@inproceedings{
sodhi2024step,
title={SteP: Stacked {LLM} Policies for Web Actions},
author={Paloma Sodhi and S.R.K Branavan and Yoav Artzi and Ryan McDonald},
booktitle={First Conference on Language Modeling},
year={2024},
url={https://openreview.net/forum?id=5fg0VtRxgi}
}

@inproceedings{besta2024graph,
  title={Graph of thoughts: Solving elaborate problems with large language models},
  author={Besta, Maciej and Blach, Nils and Kubicek, Ales and Gerstenberger, Robert and Podstawski, Michal and Gianinazzi, Lukas and Gajda, Joanna and Lehmann, Tomasz and Niewiadomski, Hubert and Nyczyk, Piotr and others},
  booktitle={Proceedings of the AAAI conference on artificial intelligence},
  volume={38},
  pages={17682--17690},
  year={2024}
}

@inproceedings{liu2025odyssey,
  title={ODYSSEY: empowering minecraft agents with open-world skills},
  author={Liu, Shunyu and Li, Yaoru and Zhang, Kongcheng and Cui, Zhenyu and Fang, Wenkai and Zheng, Yuxuan and Zheng, Tongya and Song, Mingli},
  booktitle={Proceedings of the Thirty-Fourth International Joint Conference on Artificial Intelligence},
  pages={187--195},
  year={2025}
}

@incollection{tulving1972episodic,
  title={Episodic and semantic memory},
  author={Tulving, Endel},
  booktitle={Organization of Memory},
  editor={Tulving, Endel and Donaldson, Wayne},
  pages={381--403},
  year={1972},
  publisher={Academic Press},
  address={New York, NY, USA}
}

@article{LEUNG2026115706,
title = {From entity-centric to goal-oriented graphs: Enhancing LLM knowledge retrieval in minecraft},
journal = {Knowledge-Based Systems},
volume = {341},
pages = {115706},
year = {2026},
issn = {0950-7051},
doi = {https://doi.org/10.1016/j.knosys.2026.115706},
url = {https://www.sciencedirect.com/science/article/pii/S0950705126004466},
author = {Jonathan Leung and Yongjie Wang and Zhiqi Shen}
}

@article{lake2017building,
  title={Building machines that learn and think like people},
  author={Lake, Brenden M and Ullman, Tomer D and Tenenbaum, Joshua B and Gershman, Samuel J},
  journal={Behavioral and brain sciences},
  volume={40},
  pages={e253},
  year={2017},
  publisher={Cambridge University Press}
}

@article{fang2025memp,
  title={Memp: Exploring agent procedural memory},
  author={Fang, Runnan and Liang, Yuan and Wang, Xiaobin and Wu, Jialong and Qiao, Shuofei and Xie, Pengjun and Huang, Fei and Chen, Huajun and Zhang, Ningyu},
  journal={arXiv preprint arXiv:2508.06433},
  year={2025}
}

@inproceedings{ruis2025procedural,
  title={Procedural knowledge in pretraining drives reasoning in large language models},
  author={Ruis, Laura and Mozes, Maximilian and Bae, Juhan and Kamalakara, Siddhartha Rao and Talupuru, Dwarak and Locatelli, Acyr and Kirk, Robert and Rockt{\"a}schel, Tim and Grefenstette, Edward and Bartolo, Max},
  booktitle={International Conference on Learning Representations},
  volume={2025},
  pages={29367--29429},
  year={2025}
}

@article{ouyang2025reasoningbank,
  title={Reasoningbank: Scaling agent self-evolving with reasoning memory},
  author={Ouyang, Siru and Yan, Jun and Hsu, I and Chen, Yanfei and Jiang, Ke and Wang, Zifeng and Han, Rujun and Le, Long T and Daruki, Samira and Tang, Xiangru and others},
  journal={arXiv preprint arXiv:2509.25140},
  year={2025}
}

@article{garcez2023neurosymbolic,
  title={Neurosymbolic AI: The 3rd wave},
  author={Garcez, Artur d’Avila and Lamb, Luis C},
  journal={Artificial Intelligence Review},
  volume={56},
  number={11},
  pages={12387--12406},
  year={2023},
  publisher={Springer}
}

@article{manhaeve2018deepproblog,
  title={Deepproblog: Neural probabilistic logic programming},
  author={Manhaeve, Robin and Dumancic, Sebastijan and Kimmig, Angelika and Demeester, Thomas and De Raedt, Luc},
  journal={Advances in neural information processing systems},
  volume={31},
  year={2018}
}

@book{ghallab2004automated,
  title={Automated Planning: theory and practice},
  author={Ghallab, Malik and Nau, Dana and Traverso, Paolo},
  year={2004},
  publisher={Elsevier}
}

@article{fox2003pddl2,
  title={PDDL2. 1: An extension to PDDL for expressing temporal planning domains},
  author={Fox, Maria and Long, Derek},
  journal={Journal of artificial intelligence research},
  volume={20},
  pages={61--124},
  year={2003}
}

@book{newell1994unified,
  title={Unified theories of cognition},
  author={Newell, Allen},
  year={1990},
  publisher={Harvard University Press},
  address={Cambridge, MA, USA}
}

@book{anderson2013architecture,
  title={The architecture of cognition},
  author={Anderson, John R},
  year={2013},
  publisher={Psychology Press}
}

@incollection{minsky1974framework,
  title={A framework for representing knowledge},
  author={Minsky, Marvin},
  booktitle={The Psychology of Computer Vision},
  editor={Winston, Patrick Henry},
  pages={211--277},
  year={1975},
  publisher={McGraw-Hill},
  address={New York, NY, USA}
}

@article{hafner2023mastering,
  title={Mastering diverse domains through world models},
  author={Hafner, Danijar and Pasukonis, Jurgis and Ba, Jimmy and Lillicrap, Timothy},
  journal={arXiv preprint arXiv:2301.04104},
  year={2023}
}

@article{pan2024unifying,
  title={Unifying large language models and knowledge graphs: A roadmap},
  author={Pan, Shirui and Luo, Linhao and Wang, Yufei and Chen, Chen and Wang, Jiapu and Wu, Xindong},
  journal={IEEE Transactions on Knowledge and Data Engineering},
  volume={36},
  number={7},
  pages={3580--3599},
  year={2024},
  publisher={IEEE}
}

@article{schick2023toolformer,
  title={Toolformer: Language models can teach themselves to use tools},
  author={Schick, Timo and Dwivedi-Yu, Jane and Dess{\`\i}, Roberto and Raileanu, Roberta and Lomeli, Maria and Hambro, Eric and Zettlemoyer, Luke and Cancedda, Nicola and Scialom, Thomas},
  journal={Advances in neural information processing systems},
  volume={36},
  pages={68539--68551},
  year={2023}
}

@article{ha2018world,
  title={World Models},
  author={Ha, David and Schmidhuber, J{\"u}rgen},
  journal={arXiv preprint arXiv:1803.10122},
  year={2018}
}

@article{yao2022webshop,
  title={Webshop: Towards scalable real-world web interaction with grounded language agents},
  author={Yao, Shunyu and Chen, Howard and Yang, John and Narasimhan, Karthik},
  journal={Advances in Neural Information Processing Systems},
  volume={35},
  pages={20744--20757},
  year={2022}
}

@inproceedings{qin2024toolllm,
  title={Toolllm: Facilitating large language models to master 16000+ real-world apis},
  author={Qin, Yujia and Liang, Shihao and Ye, Yining and Zhu, Kunlun and Yan, Lan and Lu, Yaxi and Lin, Yankai and Cong, Xin and Tang, Xiangru and Qian, Bill and others},
  booktitle={International Conference on Learning Representations},
  volume={2024},
  pages={9695--9717},
  year={2024}
}

@article{mcclelland1995there,
  title={Why there are complementary learning systems in the hippocampus and neocortex: insights from the successes and failures of connectionist models of learning and memory.},
  author={McClelland, James L and McNaughton, Bruce L and O'Reilly, Randall C},
  journal={Psychological review},
  volume={102},
  number={3},
  pages={419},
  year={1995},
  publisher={American Psychological Association}
}

@inproceedings{lightman2024let,
  title={Let's verify step by step},
  author={Lightman, Hunter and Kosaraju, Vineet and Burda, Yuri and Edwards, Harrison and Baker, Bowen and Lee, Teddy and Leike, Jan and Schulman, John and Sutskever, Ilya and Cobbe, Karl},
  booktitle={International Conference on Learning Representations},
  volume={2024},
  pages={39578--39601},
  year={2024}
}

@book{kahneman2011thinking,
  title={Thinking, fast and slow},
  author={Kahneman, Daniel},
  year={2011},
  publisher={Farrar, Straus and Giroux},
  address={New York, NY, USA}
}

@inproceedings{
  shridhar2021alfworld,
  title={ALFWorld: Aligning Text and Embodied Environments for Interactive Learning},
  author={Mohit Shridhar and Xingdi Yuan and Marc-Alexandre Cote and Yonatan Bisk and Adam  Trischler and Matthew Hausknecht},
  booktitle={International Conference on Learning Representations},
  year={2021},
  url={https://openreview.net/forum?id=0IOX0YcCdTn}
}

@inproceedings{jimenez2024swe,
  title={Swe-bench: Can language models resolve real-world github issues?},
  author={Jimenez, Carlos E and Yang, John and Wettig, Alexander and Yao, Shunyu and Pei, Kexin and Press, Ofir and Narasimhan, Karthik},
  booktitle={International Conference on Learning Representations},
  volume={2024},
  pages={54107--54157},
  year={2024}
}

@inproceedings{erol1994htn,
  title={HTN planning: Complexity and expressivity},
  author={Erol, Kutluhan and Hendler, James and Nau, Dana S},
  booktitle={AAAI},
  volume={94},
  pages={1123--1128},
  year={1994}
}

@article{li2024optimus,
  title={Optimus-1: Hybrid multimodal memory empowered agents excel in long-horizon tasks},
  author={Li, Zaijing and Xie, Yuquan and Shao, Rui and Chen, Gongwei and Jiang, Dongmei and Nie, Liqiang},
  journal={Advances in neural information processing systems},
  volume={37},
  pages={49881--49913},
  year={2024}
}

@article{ou2025symbolic,
  title={Symbolic learning enables self-evolving agents},
  author={Ou, Yixin and Zhou, Wangchunshu and Ding, Shengwei and Li, Long and Wu, Jialong and Wang, Tiannan and Chen, Jiamin and Wang, Shuai and Xu, Xiaohua and Zhang, Ningyu and Chen, Huajun and Jiang, Yuchen Eleanor},
  journal={AI Open},
  volume={6},
  pages={314--322},
  year={2025},
  publisher={Elsevier},
  doi={10.1016/j.aiopen.2025.11.004}
}

@article{harnad1990symbol,
  title={The symbol grounding problem},
  author={Harnad, Stevan},
  journal={Physica D: Nonlinear Phenomena},
  volume={42},
  number={1--3},
  pages={335--346},
  year={1990},
  publisher={Elsevier}
}

@incollection{hinton1986distributed,
  title={Distributed Representations},
  author={Hinton, Geoffrey E and McClelland, James L and Rumelhart, David E},
  editor={Rumelhart, David E and McClelland, James L},
  booktitle={Parallel Distributed Processing, Volume 1: Explorations in the Microstructure of Cognition: Foundations},
  year={1986},
  publisher={MIT Press},
  address={Cambridge, MA, USA}
}

@article{smolensky1990tensor,
  title={Tensor product variable binding and the representation of symbolic structures in connectionist systems},
  author={Smolensky, Paul},
  journal={Artificial intelligence},
  volume={46},
  number={1-2},
  pages={159--216},
  year={1990},
  publisher={Elsevier}
}

@book{kanerva1988sparse,
  title={Sparse Distributed Memory},
  author={Kanerva, Pentti},
  year={1988},
  publisher={MIT Press},
  address={Cambridge, MA, USA}
}

@article{plate1995hrr,
  title={Holographic reduced representations},
  author={Plate, Tony A},
  journal={IEEE Transactions on Neural Networks},
  volume={6},
  number={3},
  pages={623--641},
  year={1995},
  publisher={IEEE}
}

@article{richardson2006markov,
  title={Markov logic networks},
  author={Richardson, Matthew and Domingos, Pedro},
  journal={Machine learning},
  volume={62},
  number={1},
  pages={107--136},
  year={2006},
  publisher={Springer}
}

@article{ellis2023dreamcoder,
    author = {Ellis, Kevin and Wong, Lionel and Nye, Maxwell and Sablé-Meyer, Mathias and Cary, Luc and Anaya Pozo, Lore and Hewitt, Luke and Solar-Lezama, Armando and Tenenbaum, Joshua B.},
    title = {DreamCoder: growing generalizable, interpretable knowledge with wake–sleep Bayesian program learning},
    journal = {Philosophical Transactions of the Royal Society A: Mathematical, Physical and Engineering Sciences},
    volume = {381},
    number = {2251},
    pages = {20220050},
    year = {2023},
    month = {06},
    issn = {1364-503X},
    doi = {10.1098/rsta.2022.0050},
    url = {https://doi.org/10.1098/rsta.2022.0050},
    eprint = {https://royalsocietypublishing.org/rsta/article-pdf/doi/10.1098/rsta.2022.0050/1327751/rsta.2022.0050.pdf},
}

@book{hayesroth1983expert,
  title={Building Expert Systems},
  editor={Hayes-Roth, Frederick and Waterman, Donald A and Lenat, Douglas B},
  year={1983},
  publisher={Addison-Wesley},
  address={Reading, MA, USA}
}

@article{cullen1988reassessment,
  title={The knowledge acquisition bottleneck: time for reassessment?},
  author={Cullen, John and Bryman, Alan},
  journal={Expert Systems},
  volume={5},
  number={3},
  pages={216--225},
  year={1988},
  publisher={Wiley},
  doi={10.1111/j.1468-0394.1988.tb00065.x}
}

@article{kirkpatrick2017overcoming,
  title={Overcoming catastrophic forgetting in neural networks},
  author={Kirkpatrick, James and Pascanu, Razvan and Rabinowitz, Neil and Veness, Joel and Desjardins, Guillaume and Rusu, Andrei A and Milan, Kieran and Quan, John and Ramalho, Tiago and Grabska-Barwinska, Agnieszka and others},
  journal={Proceedings of the National Academy of Sciences},
  volume={114},
  number={13},
  pages={3521--3526},
  year={2017},
  publisher={National Academy of Sciences}
}

@article{frankland2005organization,
  title={The organization of recent and remote memories},
  author={Frankland, Paul W and Bontempi, Bruno},
  journal={Nature Reviews Neuroscience},
  volume={6},
  number={2},
  pages={119--130},
  year={2005},
  publisher={Nature Publishing Group}
}

@article{singhal2023large,
  title={Large language models encode clinical knowledge},
  author={Singhal, Karan and Azizi, Shekoofeh and Tu, Tao and others},
  journal={Nature},
  volume={620},
  number={7972},
  pages={172--180},
  year={2023},
  publisher={Springer Nature}
}

@article{kaelbling1998planning,
  title={Planning and acting in partially observable stochastic domains},
  author={Kaelbling, Leslie Pack and Littman, Michael L and Cassandra, Anthony R},
  journal={Artificial Intelligence},
  volume={101},
  number={1--2},
  pages={99--134},
  year={1998},
  publisher={Elsevier}
}

@article{ye2024moreagents,
  title={More Agents Is All You Need},
  author={Li, Junyou and Zhang, Qin and Yu, Yangbin and Fu, Qiang and Ye, Deheng},
  journal={Transactions on Machine Learning Research},
  year={2024}
}

@inproceedings{ye2024avalon,
  title={LLM-based Agent Society Investigation: Collaboration and Confrontation in Avalon Gameplay},
  author={Lan, Yihuai and Hu, Zhiqiang and Wang, Lei and Wang, Yang and Ye, Deheng and Zhao, Peilin and Lim, Ee-Peng and Xiong, Hui and Wang, Hao},
  booktitle={Proceedings of the 2024 Conference on Empirical Methods in Natural Language Processing (EMNLP)},
  year={2024}
}

@article{ye2024affordable,
  title={Affordable Generative Agents},
  author={Yu, Yangbin and Zhang, Qin and Li, Junyou and Fu, Qiang and Ye, Deheng},
  journal={Transactions on Machine Learning Research},
  year={2024}
}

@article{ye2023rltf,
  title={RLTF: Reinforcement Learning from Unit Test Feedback},
  author={Liu, Jiate and Zhu, Yiqin and Xiao, Kaiwen and Fu, Qiang and Han, Xiao and Yang, Wei and Ye, Deheng},
  journal={Transactions on Machine Learning Research},
  year={2023}
}

@inproceedings{ye2023fcdt,
  title={Future-conditioned Unsupervised Pretraining for Decision Transformer},
  author={Xie, Zhihui and Lin, Zichuan and Ye, Deheng and Fu, Qiang and Wei, Yang and Li, Shuai},
  booktitle={International Conference on Machine Learning (ICML)},
  pages={38187--38203},
  year={2023},
  organization={PMLR}
}

@article{ye2024walle,
  title={WALL-E: World Alignment by Rule Learning Improves World-Model-Based LLM Agents},
  author={Zhou, Siyu and Zhou, Tianyi and Yang, Yijun and Long, Guodong and Ye, Deheng and Jiang, Jing and Zhang, Chengqi},
  journal={arXiv preprint arXiv:2410.07484},
  year={2024}
}

@inproceedings{ye2021mapgo,
  title={MapGo: Model-Assisted Policy Optimization for Goal-Oriented Tasks},
  author={Zhu, Menghui and Liu, Minghuan and Shen, Jian and Zhang, Zhicheng and Chen, Sheng and Zhang, Weinan and Ye, Deheng and Yu, Yong and Fu, Qiang and Yang, Wei},
  booktitle={Proceedings of the Thirtieth International Joint Conference on Artificial Intelligence (IJCAI)},
  year={2021}
}

@inproceedings{ye2021juewumc,
  title={Juewu-MC: Playing Minecraft with Sample-Efficient Hierarchical Reinforcement Learning},
  author={Lin, Zichuan and Li, Junyou and Shi, Jianing and Ye, Deheng and Fu, Qiang and Yang, Wei},
  booktitle={Proceedings of the Thirtieth International Joint Conference on Artificial Intelligence (IJCAI)},
  year={2021}
}

@article{ye2021tikick,
  title={TiKick: Towards Playing Multi-Agent Football Full Games from Single-Agent Demonstrations},
  author={Huang, Shiyu and Chen, Wenze and Zhang, Longfei and Xu, Shizhen and Li, Ziyang and Zhu, Fengming and Ye, Deheng and Chen, Ting and Zhu, Jun},
  journal={arXiv preprint arXiv:2110.04507},
  year={2021}
}

@inproceedings{ye2026wallens,
  title={WALL-E: World Alignment by Neurosymbolic Learning Improves World-Model-Based LLM Agents},
  author={Zhou, Siyu and Zhou, Tianyi and Yang, Yijun and Long, Guodong and Ye, Deheng and Jiang, Jing and Zhang, Chengqi},
  booktitle={Advances in Neural Information Processing Systems},
  volume={38},
  pages={65352--65391},
  year={2026}
}

@inproceedings{ye2025vistawise,
  title={Vistawise: Building Cost-Effective Agent with Cross-Modal Knowledge Graph for Minecraft},
  author={Fu, Haobo and Ren, Jiaqi and Chai, Qiang and Ye, Deheng and Cai, Yiyu and Wang, Hao},
  booktitle={Proceedings of the 2025 Conference on Empirical Methods in Natural Language Processing (EMNLP)},
  year={2025}
}

@inproceedings{ye2025causalmace,
  title={CausalMACE: Causality Empowered Multi-Agents in Minecraft Cooperative Tasks},
  author={Chai, Qiang and Zheng, Zirui and Ren, Jiaqi and Ye, Deheng and Lin, Zichuan and Wang, Hao},
  booktitle={Findings of the Association for Computational Linguistics: EMNLP 2025},
  pages={14410--14426},
  year={2025}
}

@article{ye2026uivoyager,
  title={Ui-Voyager: A Self-Evolving GUI Agent Learning via Failed Experience},
  author={Lin, Zichuan and Liu, Feng and Yang, Yijun and Lyu, Jiafei and Gao, Yifei and Liu, Yang and Lu, Zongqing and Yu, Yangbin and Yang, Meng and Li, Junyou and Ye, Deheng},
  journal={arXiv preprint arXiv:2603.24533},
  year={2026}
}

@inproceedings{ye2022minerl,
  title={MineRL Diamond 2021 Competition: Overview, Results, and Lessons Learned},
  author={Kanervisto, Anssi and Milani, Stephanie and Ramanauskas, Karolis and Topin, Nicholay and Lin, Zichuan and Li, Junyou and Shi, Jianing and Ye, Deheng and Fu, Qiang and Yang, Wei and Hong, Weijun and Huang, Zhongyue and Chen, Haicheng and Zeng, Guangjun and Lin, Yue and Micheli, Vincent and Alonso, Eloi and Fleuret, Fran{\c{c}}ois and Nikulin, Alexander and Belousov, Yury and Svidchenko, Oleg and Shpilman, Aleksei},
  booktitle={NeurIPS 2021 Competitions and Demonstrations Track},
  pages={13--28},
  year={2022},
  organization={PMLR}
}

@inproceedings{ye2021uismells,
  title={Don't Do That! Hunting Down Visual Design Smells in Complex UIs Against Design Guidelines},
  author={Yang, Bo and Xing, Zhenchang and Xia, Xin and Chen, Chunyang and Ye, Deheng and Li, Shanping},
  booktitle={2021 IEEE/ACM 43rd International Conference on Software Engineering (ICSE)},
  pages={761--772},
  year={2021}
}

@article{ye2023tfrlsurvey,
  title={A Survey on Transformers in Reinforcement Learning},
  author={Li, Wenzhe and Luo, Hao and Lin, Zichuan and Zhang, Chongjie and Lu, Zongqing and Ye, Deheng},
  journal={Transactions on Machine Learning Research},
  year={2023}
}

@article{ye2022pretrainrl,
  title={Pretraining in Deep Reinforcement Learning: A Survey},
  author={Xie, Zhihui and Lin, Zichuan and Li, Junyou and Li, Shuai and Ye, Deheng},
  journal={arXiv preprint arXiv:2211.03959},
  year={2022}
}

@inproceedings{ye2021relationtransformer,
  title={Relation-aware Transformer for Portfolio Policy Learning},
  author={Xu, Ke and Zhang, Yifan and Ye, Deheng and Zhao, Peilin and Tan, Mingkui},
  booktitle={Proceedings of the Thirtieth International Joint Conference on Artificial Intelligence (IJCAI)},
  pages={4647--4653},
  year={2021}
}

@inproceedings{ye2025crossdomain,
  title={Cross-domain Offline Policy Adaptation with Optimal Transport and Dataset Constraint},
  author={Lyu, Jiafei and Yan, Mengbei and Qiao, Zhongjian and Liu, Runze and Ma, Xiaoteng and Ye, Deheng and Yang, Jing-Wen and Lu, Zongqing and Li, Xiu},
  booktitle={International Conference on Learning Representations (ICLR)},
  pages={85653--85682},
  year={2025}
}

\section*{Figure legends}

\textbf{Figure 1.} \textbf{Procedural memory as a distinct memory system in LLM-based agents.} The LLM serves as a reasoning core containing implicit procedural knowledge alongside an agent cognition layer integrating three complementary external memory components. Semantic memory (RAG, knowledge graphs) and episodic memory (experience replay, reflection) provide factual and experiential context. Procedural memory encodes structured knowledge of how tasks are accomplished and is the focus of this perspective. It may be represented implicitly within LLM parameters or externally through experience-based procedures and symbolic goal structures, which are shown as complementary procedural memory representations. Within procedural memory, experience-based procedures may learn incrementally from interaction (gray arrow), while symbolic goal structures (GoG, Goal Net) currently rely on offline construction, with online refinement from experience as an open direction (dotted arrow).

\textbf{Figure 2.} \textbf{Illustrative example of Goal-Oriented Graphs as explicit procedural memory.} \textit{Left}: a partial subgraph for the task ``craft a diamond axe''. Nodes represent goals; edges encode subgoal dependencies. Shared dependencies, such as ``craft sticks'', required by both the axe and the iron pickaxe, make the structure a graph rather than a tree, enabling procedural knowledge to be reused across task branches. Shaded nodes are atomic operations. Not all goals and subgoals are shown; the full goal graph contains additional nodes including tool prerequisites such as the crafting table. \textit{Right}: the retrieval-reasoning loop. Given a user goal, the relevant subgraph is retrieved via embedding search, serialised into natural language, and injected into the LLM context to guide plan generation. When execution fails, the GoG provides the prerequisite chain for the missing resource, enabling targeted replanning rather than complete plan regeneration~\cite{LEUNG2026115706}.

\end{document}